\documentclass[review,sort&compress,margin=1.3in]{elsarticle}

\usepackage{float}

\usepackage{stfloats}

\usepackage{hyperref} 
\hypersetup{
    colorlinks=true
}

\usepackage[labelfont=bf]{caption}

\usepackage[none]{hyphenat}  

\usepackage{multirow}

\usepackage{tabularx} 

\usepackage[switch]{lineno}

\usepackage{xcolor}

\usepackage[british]{babel}

\usepackage{xstring}
\makeatletter
\AtBeginDocument{
\let\oldref\ref
\renewcommand{\ref}[1]{\IfBeginWith{#1}{fig:}%
{{\color{blue}Figure~\oldref{#1}}}%
{\IfBeginWith{#1}{tab:}{{\color{blue}Table~\oldref{#1}}}{Unsupported ref start}}}%
}
\makeatother

\begin{document}

\title{Cross-Resolution Learning for Face Recognition}

\begin{frontmatter}

\author[address]{Fabio Valerio Massoli\corref{cor1}}
\ead{fabio.massoli@isti.cnr.it}

\author[address]{Giuseppe Amato}
\author[address]{Fabrizio Falchi}

\cortext[cor1]{Corresponding author}
\address[address]{ISTI-CNR, via G. Moruzzi 1, 56124 Pisa, Italy}

\begin{abstract}

Convolutional Neural Networks have reached extremely high performances on the Face Recognition task.
Largely used datasets, such as VGGFace2, focus on gender, pose and age variations trying to balance them in order to achieve better results. However, the fact that images have different resolutions is not usually discussed and resize to 256 pixels before cropping is used. While specific datasets for very low resolution faces have been proposed, less attention has been payed on the task of cross-resolution matching.
Such scenarios are of particular interest for forensic applications and surveillance systems in which it usually happens that a low-resolution probe has to be matched with higher-resolution galleries.
While it is always possible to either increase the resolution of the probe image or to reduce the size of the gallery images (less frequently), to the best of our knowledge an extensive experimentation of cross-resolution matching was missing in the recent deep learning based literature.
In the context of low- and cross-resolution Face Recognition, the contribution of our work is fourfold: 
i) we proposed a training method to fine-tune a state-of-the-art model in order to make it able to extract resolution-robust deep features; 
ii) we tested our models on the benchmark datasets IJB-B and IJB-C considering images at both full and low resolutions in order to show the effectiveness of the proposed training algorithm. To the best of our knowledge, this is the first work testing extensively the performance of a FR model in a cross-resolution scenario; 
iii) we tested our models on the low resolution and low quality datasets QMUL-SurvFace and TinyFace and showed their superior performances, even though we did not train our model on low-resolution faces only and our main focus was cross-resolution; 
iv) we showed that our approach can be more effective with respect to preprocessing faces with super resolution techniques. \\
The python code of the proposed method will be available at \url{https://github.com/fvmassoli/gsda}.

\end{abstract}

\begin{keyword}
Deep Learning\sep Low Resolution Face Recognition\sep Cross Resolution Face Recognition
\end{keyword}

\end{frontmatter}

\section{Introduction}
Face Recognition (FR) has always been among the most relevant topics in computer vision. 
Thanks to the recent advances in neural network methods, the large availability of powerful GPUs and the creation of very large datasets, the use of Deep Learning (DL) techniques~\cite{ref_article1,ref_article2} have become state-of-the-art to solve FR tasks.
The success reached by DL models and more specifically by Deep Convolutional Neural Networks (DCNNs) is due to their peculiar multi-layer structure that allows to create inner representations of the input images which have a high discrimination power.\\
The first attempts for solving the FR problem date back to the early 90s when the so called Eigenfaces~\cite{turk1991face} approach was proposed. Subsequently, local features-based methods, such as Gabor~\cite{liu2002gabor} and LBP~\cite{ahonen2006face}, pushed the performance on the FR task forward until, starting from 2012, the DL models started to dominate the field reaching performances up to 99.80\%~\cite{wang2018deep} thus, overcoming human performance on the very same task.\\
It is generally observed that even though DL algorithms perform well when tested against images taken under controlled conditions, e.g. high-resolution (HR) and frontal pose, a sudden drop in their performance occurs when they are tested against images taken under uncontrolled conditions, e.g. low-resolution (LR). For example, this situation emerges in the context of surveillance systems~\cite{ref_article4} which typically rely on cameras with limited resolution. Thus, a cross-resolution problem arises in which a probe with variable and often very low resolution has to be matched against HR galleries making the FR a much harder task to fulfill.\\
In 2018, Cao et al.~\cite{cao2018vggface2} published a new face datasets, called VGGFace2, with which they trained a DCNN models that set the new state-of-the-art for FR tasks. Even though the dataset is characterized by a high pose, age and illumination variability, which make it a suitable choice in order to train DL models, resolution variations have not been taken into account while collecting the dataset nor while training the models. Indeed, as we will show later in the paper, also the new state-of-the-art model from~\cite{cao2018vggface2} experiences a drop in performances in the low- and an even more severe decrease in the cross-resolution scenarios.\\
Generally, the two techniques used to deal with LR images are: the Super Resolution~\cite{ref_article5,ref_article6} and the projection of the LR probe and the HR gallery into a common space~\cite{ref_article7}.\\
Apart from DL methods, other approaches have also been used in order to tackle the problem of cross-resolution FR such as methods based on the discrete wavelet transform~\cite{ekenel2005multiresolution} and on dictionary learning~\cite{luo2019multi}.\\
In this work, we consider a different approach. Indeed, we propose a training algorithm by means of which it is possible to make a DL model more robust to variation in the input images resolution. Moreover, we will show how our techniques is beneficial not only for the cross-resolution FR task but also when we consider images at the same, but very low, resolution (down to 8 pixels).
Specifically, we took our steps from the state-of-the-art ResNet-50~\cite{he2016deep} architecture, equipped with Squeeze-and-Excitation blocks~\cite{hu2017squeeze}, from Cao et al.~\cite{cao2018vggface2} and we improved upon its performances on the aforementioned tasks showing that it is possible to adapt a model to the LR domain without losing too much, in terms of performance, at HR.\\
After the training phase, we extensively tested our models on the 1:1 face verification and 1:N face identification protocols. We considered the IARPA Janus Benchmark-B (IJB-B)~\cite{whitelam2017iarpa} and the IARPA Janus Benchmark-C (IJB-C)~\cite{maze2018iarpa} benchmark datasets and we used several versions of them, one at full resolution and others at lower resolutions, in order to asses the performances of our \mbox{models}.\\
Then, we tested the robustness of our models on low resolution and low quality images benchmark datasets QMUL-SurvFace~\cite{cheng2018surveillance} and TinyFace~\cite{cheng2018low}.\\
The rest of the paper is organized as follows. In \hyperlink{motivation}{Section 2} we give some insights into the motivations for our study. \hyperlink{related_works}{Section 3} briefly presents some related works while in \hyperlink{face_datasets}{Section 4} we describe all the datasets we have used. \hyperlink{proposed_training_approach}{Section 5} gives a detailed description of the training algorithm. Finally, in \hyperlink{experimental_results}{Sections 6} and \hyperlink{conclusion}{7} we show the experimental results and reported the conclusions, respectively. 

\section{Motivation} \hypertarget{motivation}
Typically, while collecting images for FR datasets, the resolution variation is not a major concern. Perhaps, this could be due to the fact that since they generally consist of images of famous people's faces like politicians, athletes, actors etc., their pictures available on the web are often at HR. 
Examples of such datasets are MS-Celeb-1M~\cite{guo2016ms}, VGGFace~\cite{parkhi2015deep} and VGGFace2~\cite{cao2018vggface2}.\\
Thus, if the resolution variation is not properly considered while training, the FR task can be very hard to fulfill by a DL model when low- and especially cross-resolution scenarios are met.
In a first attempt to remedy to such an issue, few low-resolution datasets were published. An example are the UCCS~\cite{gunther2017unconstrained} and the ScFace~\cite{grgic2011scface} datasets. Recently, two new larger and more challenging datasets have been made publicly available: QMUL-SurvFace~\cite{cheng2018surveillance} and TinyFace~\cite{cheng2018low}, which contain LR images extracted from surveillance cameras and from the web, respectively. Unfortunately, even though they comprehend low-resolution and low-quality images, their multiplicity is about one order of magnitude lower than the previously mentioned HR datasets, and for this reason it becomes arduous to utilize them to train state-of-the-art models.\\ Thus, it is consistently observed that the performance of DL models degrades dramatically when tested against low- and cross-resolution FR tasks that are scenarios of great interest to forensic and surveillance system applications. Indeed, this is exactly what we observed for the state-of-the-art model from~\cite{cao2018vggface2}. The authors focused their efforts on gender, age and pose variations reaching the highest performances ever on the FR tasks of verification and identification. Nevertheless, since they did not consider face resolution variation in their study, as we will show later in the paper, their model suffers from a performance drop especially when tested against the cross-resolution FR task. \\
Therefore, it is mandatory to enable DL models to extract resolution-robust deep features. In this context we address this issue by smoothly adapt a HR state-of-the-art model to the LR domain by using our training algorithm. Being able to generate resolution-robust deep features, such model will achieve high performances regardless of the input image resolution. 

\section{Related works} \hypertarget{related_works}
In~\cite{ekenel2005multiresolution} the FR problem is tackled from a non-DL perspective. Starting from the original images, the authors performed a multi-resolution analysis by decomposing the face image into frequency subbands by means of 2D discrete wavelet transform. Their goal was to improve upon face recognition performances considering variations in expression and illumination. In order to achieve that, they look for the subbands less sensitive to that kind of variations. \\
Dictionary learning is a technique in which a learning model exploits some or all training images to represent this image based on a learned dictionary. Typically, in this context researchers focus only on a single resolution. In~\cite{luo2019multi} they exploit a multi-resolution dictionary learning method that provides multiple dictionaries each
being associated with a resolution. With their methods, the authors have been able to achieve comparable results, on some face databases, with the ones from deep learning models pre-trained on the ImageNet dataset. \\
Super Resolution~\cite{ref_article6} (SR, or Face Hallucination) is a procedure in which a deep model is used to synthesize a HR image starting from a LR probe. A frequent objection to this technique is that the synthesis process is not optimized for discrimination tasks. Thus, the information about the identity of the person might be lost. In~\cite{ref_article5}, the attempt is to address this issue by introducing an identity loss in order to impose the identity information during the training process. Hence, the aim is to recover the initial LR image identity in the HR one.\\
In~\cite{zou2011very} the authors tackled the LR FR problem by formulating an approach based on learning a relationship between the HR and the LR image spaces. In their study, they adopted two constraints while learning the relationship: one for the visual quality of the reconstructed images and another in order to generate discriminative images from a machine point of view. The learned relationship is then applied at inference time in order to synthetize HR images from LR ones.\\
~\cite{tb} introduces a two-branches model which learns nonlinear transformations in order to map high- and low-resolution images into a common space. One branch accepts the HR images as inputs, while the other allows the LR ones. Both branches produce feature vectors that are then projected into a common space where their distance is evaluated. The error on the distance is then backprogated only through the bottom branch with the goal of minimizing the distance between features vectors of HR and LR images.\\
In~\cite{sb}, Shekhar et al. tackled the LR FR problem by means of a generative technique rather than a discriminative one, building their approach on learning class specific dictionaries that, moreover, resulted to be robust with respect to illumination variations.\\
In~\cite{lu2018deep} the authors used a CNN-based approach in order to learn a mapping to a common space for high- and low-resolution images. The overall architecture is divided into two sections: a trunk network for features extraction and a two branch network which learns a coupled mapping which project HR and LR vectors to a common space where their distance is minimized. The trunk network is only trained once with images at three different resolutions. Subsequently, several two-branches networks are trained for different image resolutions.\\

\section{Face datasets} \hypertarget{face_datasets}
In this Section we provide an overview of the datasets used in our experiments. VGGFace2 is the only dataset we used in order to train our models, whereas all the others were adopted for test purposes only.

\subsection{VGGFace2}
\sloppy
VGGFace2~\cite{cao2018vggface2} is made of $\sim$3.31 million images divided in 9131 classes, each representing a different person identity. The dataset is divided into two splits, one for the training and one for test. The latter contains $\sim$170000 images divided into 500 identities while all the other images belong to the remaining 8631 classes available for training. While constructing the datasets, the authors focused their efforts on reaching a very low label noise and a high pose and age diversity thus, making the VGGFace2 dataset a suitable choice to train state-of-the-art deep learning models on face-related tasks.\\
Nevertheless, the problem of resolution variations has not been taken into account. Indeed, the images of the training set have an average resolution of $\sim$137x180 pixels, with less than 1\% at a resolution below 32 pixels (considering the shortest side). This can set a severe limitation on the models performances when tested in the low- and in the even harder cross-resolution scenarios. Indeed, even though Cao et al.~\cite{cao2018vggface2} have been able to train a DCNN which set the state-of-the-art for FR tasks at HR, we will show later in the paper that their model suffer from a performance drop when tested against low- and cross-resolution FR tasks. 
Thus, in order to train a resolution-robust DL model, it is mandatory to properly design a training algorithm which takes into account variations in the input image resolution.

\subsection{IJB-B}
\sloppy
The goal of the IARPA's Janus program~\cite{iarpa} is to improve upon the current performance of face recognition tools by fusing the rich spatial, temporal, and contextual information available from the multiple views captured by today's ``media in the wild". The program comprises datasets that can be used in order to benchmark face detection and recognition models on several different tasks.\\
The IARPA Janus Benchmark-B~\cite{whitelam2017iarpa} (IJB-B) face dataset is designed especially for test purposes. It consists of $\sim$76K images, from still images and video, that are shared among 1845 identities. The dataset is characterized by human-labeled ground truth face bounding boxes, eye/nose location, and covariate metadata such as occlusion, facial hair, and skin tone. Test protocols for this dataset represent operational use cases including access point identification, forensic quality media searches, surveillance video searches, and clustering. Specifically, we used the 1:1 verification and 1:N identification protocols as specified in the IARPA Janus program.\\
In order to evaluate the face verification protocol, it is provided a list of comparisons between two galleries templates, S1 and S2, and mixed media probe templates.\\ 
The 1:N identification protocol tests the accuracy of both open- and close-set identifications using probe templates from all subjects from the two disjoint galleries, S1 and S2. Each gallery set contains a single template per subject that is created by using half of the still imagery media, randomly chosen for that subject. The remaining images are reserved for the probe set for each subject.

\subsection{IJB-C}
\sloppy
The IARPA Janus Benchmark-C~\cite{maze2018iarpa} dataset advances the goal of robust unconstrained face recognition, improving upon the IJB-B~\cite{whitelam2017iarpa} dataset, by increasing the dataset size and its variability. It comprises 3531 subjects with a total of 31.3K still images (21.3K face and 10K non-face), averaging to $\sim$6 images per subject, and 117.5K frames, averaging to $\sim$33 frames per subject. Moreover, subjects show full variation in pose and diverse occupation, thus avoiding the pitfall of ``celebrity-only" media. Similarly to IJB-B~\cite{whitelam2017iarpa}, this dataset offers two disjoint galleries, G1 and G2, and gallery and probe templates are constructed with a similar approach.\\ As we did for the IJB-B~\cite{whitelam2017iarpa} dataset, we considered once more the 1:1 verification and 1:N identification (open- and close-set) protocols.

\subsection{TinyFace}
\begin{figure}[H]
\centering 
\includegraphics[width=\linewidth]{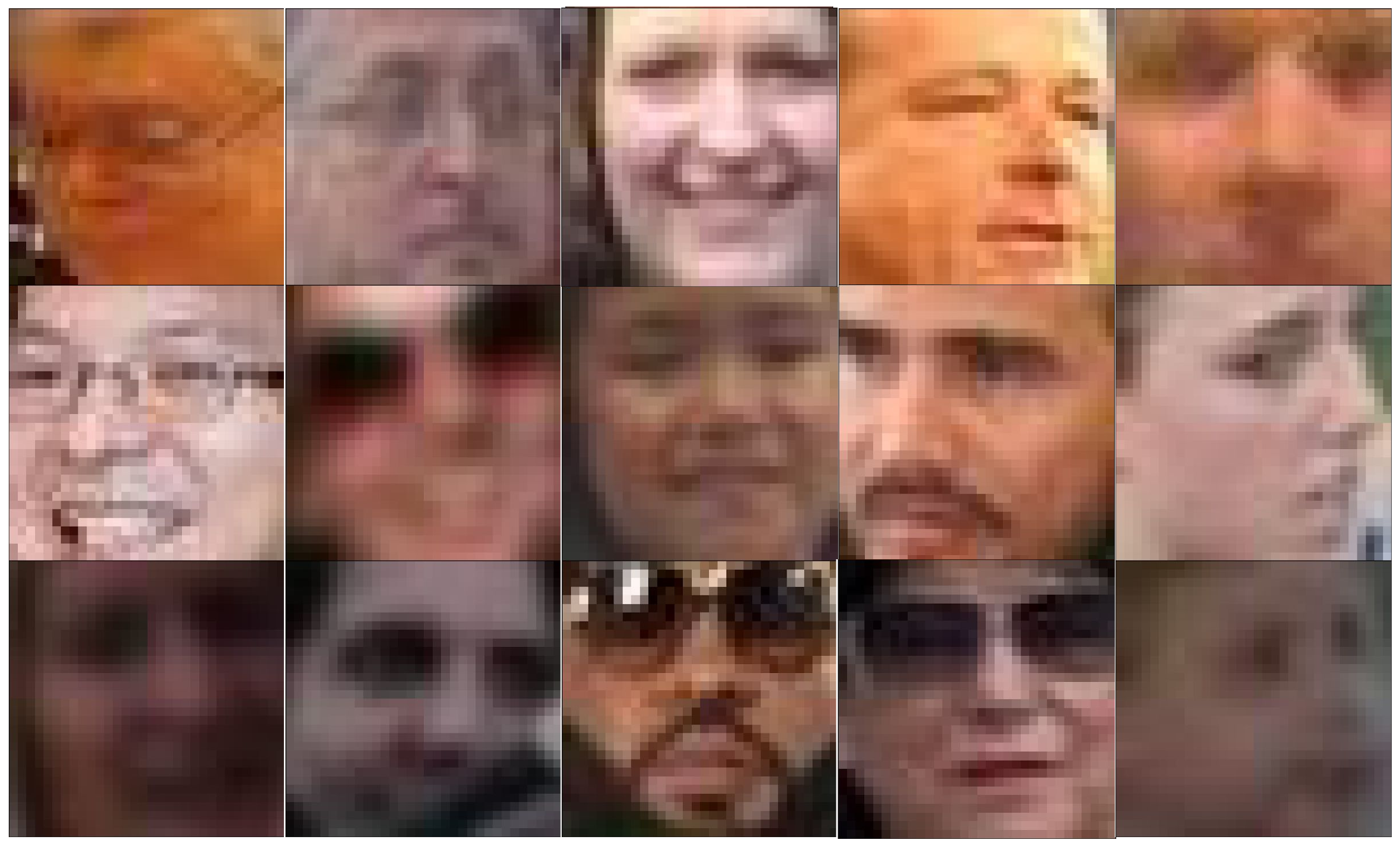}
\caption{Sample images from TinyFace~\cite{cheng2018low} dataset.} \label{fig:tinyface_example}
\end{figure}

\sloppy
The TinyFace~\cite{cheng2018low} dataset consists of 169.4K LR face images (average 20x16 pixels) designed to train and benchmark deep learning models on very LR FR tasks. It is the largest LR web face recognition benchmark. Among all the images, about 16K have distinct identity labels that represent 5139 different identities, while the others are non-labelled and used as ``distractors" for the test protocol. Images are separated into training and test splits. Specifically, 7804 images (2570 identities) are designed for training and the rest for testing. All the LR faces in TinyFace are collected from public web data across a large variety of scenarios, captured under uncontrolled viewing conditions in pose, illumination, occlusion and background. \autoref{fig:tinyface_example} shows a collection of few samples from the dataset.
We used only the test set and the given 1:N face identification (close-set) protocol. The probe set includes half of the labeled images while the other half is used to construct the gallery set to which they are subsequently added all of the non-labelled images. 

\subsection{SCface}
The SCface~\cite{grgic2011scface} dataset comprises 4160 images, that belong to 130 subjects, that were acquired in uncontrolled indoor environment using five video surveillance cameras of various qualities.\\
The images are divided into three disjoint sets on a per-subject basis: world, dev and test sets which comprises 43, 44, and 43 subjects, respectively.\\
The former two are intended to be used for training purposes while the latter is meant only for test models performances.\\
For our purposes, we only used the test set. Considering that in that face set there are 16 images for each identity, we used a total of 688 images for model tests.\\

\section{Proposed training approach} 
\begin{figure}[!h]
\centering
\includegraphics[width=\textwidth]{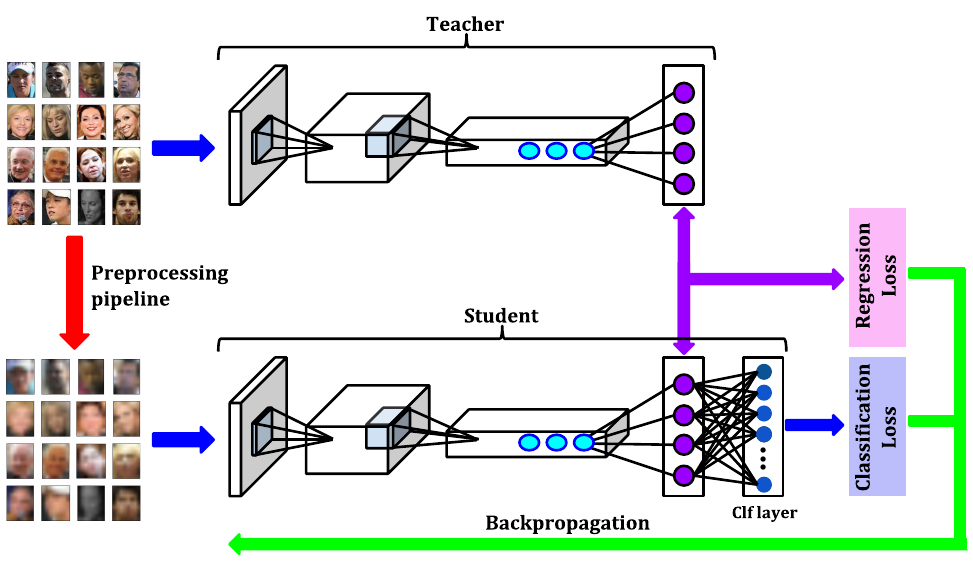}
\caption{Schematic view of the training algorithm. The Teacher weights are frozen and the gradients only flow through the Student model.} \label{fig:sketch}
\end{figure}

\hypertarget{proposed_training_approach}
The final goal of our training procedure was to create a model that is able to extract deep features that are robust with respect to variation in the input image resolution. Even though we mainly focused on the cross-resolution scenario in which a LR probe has to be matched against a HR gallery, which is a typical scenario in forensic applications and surveillance systems, we obtained noticeable improvements also in the case of FR considering images at the same, but low, resolution i.e. below 64 pixels down to 8 pixels.\\
Our starting point has been the SeNet-50 architecture from~\cite{cao2018vggface2}, published in 2018, with which the authors set the state-of-the-art for the 1:1 face verification and 1:N face identification (Tables VI, VII and VIII, penultimate row) on the IARPA's Janus~\cite{iarpa} datasets. With their best model, the authors reached, for example, a TAR value, at FAR=$1e^{-3}$, of 0.908 on the IJB-B~\cite{whitelam2017iarpa} and of 0.927 on the IJB-C~\cite{maze2018iarpa} datasets, respectively.\\
As we will show in \hyperlink{experimental_results}{Section 6}, it is worth noticing that even though the model set the state-of-the-art, its performance degraded considerably when considering images with resolutions below 64 pixels.\\
Before the final formulation of our approach, we conducted several experimentation. 
We started by training the SeNet-50 architecture from scratch using the VGGFace2~\cite{cao2018vggface2} dataset. Since we were interested in enabling the model to generate resolution-robust deep features, while training we fed the it with images at original as well as at low resolution by random down sampling them, at a fixed frequency, between 8 and 256 pixels. Unfortunately, with this strategy we obtained models with very low performances. Perhaps, this was due to the fact that LR images carries less information w.r.t. HR ones thus, making it harder for a model to converge to a good minimum. 
As a second step we decided to fine-tuned the state-of-the-art SeNet-50 model from~\cite{cao2018vggface2}. We first tried by freezing the entire net and let only the final FC layer to be trained. Then we ran new experiments by freeing all the net parameters. In both cases we fed the models with variable resolution images. We measured higher performances in the latter case and this could be due to the fact that there were patterns in the LR domain that the models needed to adjust their parameters for.\\ 
Apart from the improvements on the cross-resolution FR task up to an image resolution of 24 pixels (shortest side), we noticed a relevant drop at higher resolutions. In order to reduce such a drop we tried to smoothly adapt the model weights to the LR domain. In order to accomplish the task, we borrowed from the curriculum learning paradigm~\cite{bengio2009curriculum} the concept that a model can benefit from a training strategy in which more complex data are given as inputs as the learning process proceeds. In addition, we used the Teacher-Student~\cite{hinton2015distilling} scheme. A schematic view of the training algorithm is shown in \autoref{fig:sketch}.
As previously mentioned, we used the VGGFace2~\cite{cao2018vggface2} dataset for training. 
At the beginning, the Teacher and the Student were both the state-of-the-art SeNet-50 from~\cite{cao2018vggface2}. The difference among them is that the first one had its weights frozen and it was used only for features extraction. During the training, the Teacher was fed with full resolution images while the Student received, as inputs, variable resolution ones. The underlying idea was that by extracting the deep features from both models and pointwise comparing them, we should have enabled the Student model to approximate the same representation created by the Teacher, using HR images, with inputs of any resolution. Thus, the Student should learn to construct more resolution-robust deep representations.\\
We used the curriculum learning paradigm in order to give the Student images at low- and high-resolution as inputs. Specifically, the frequency at which a down sampled version of an image, instead of its full resolution one, was fed to the network under training, increased in a linear way from 0 to 1 while the training proceeded. The resolution at which each image was down sampled was randomly extracted in the range of [8, 256] pixels. Precisely, we uniformly extracted an integer between 3 and 8 and used it as the exponent of a power of two which indicated the final input resolution.\\ 
To be more detailed, the images were first down sampled so that the shortest side was equal to the extracted resolution while keeping the original aspect ratio, and then they were resized at the original dimensions by means of the bilinear interpolation algorithm from the PIL python library.\\
After this first preprocessing phase, the input images were resized so that the size of the shortest side was 256 pixels, then a random crop was applied in order to select a 224x224 pixel region which matched the input of the network.\\
In order to evaluate the loss, we combined the softmax loss for classification and the MSE among the deep features, extracted at the penultimate layer of the CNNs. The overall loss is reported in  \autoref{eq:loss}.\\
\begin{equation}\label{eq:loss}
    \mathcal{L} = \mathcal{L}(y, \hat{y}) + \lambda \cdot \sum_{i, i' \in \mathcal{I}} \parallel \mathcal{F_T}_{(i)} - \mathcal{F_S}_{(i')} \parallel ^{2}_{2}
\end{equation}
In \autoref{eq:loss}, the $\lambda$ term is a balancing weight. We empirically found that the best value for it was 0.1. The sum runs over the input images and $i'$ represents the down sampled version of the image $i$. Thus, the second term of the loss forces the Student model to learn a features representation for each image, independently of its resolution, which is as close as possible to the one produced from HR images. We later used a batch size of 256, the SGD optimizer with momentum equals to 0.9 and a weight decay of 1e$^{-5}$.\\
We set the initial value of the learning rate at $1.e^{-3}$ and then we decreased it by a factor 5 every time the loss reached a plateau.\\
While training we needed to monitor the performances of our models on both low- and high-resolution images. In order to accomplish it, we split the dataset into training set and validation set and then we used two versions of the latter split. One one version, we down sampled all the images to a reference resolution of 24 pixels and on the other one we used them at full resolution.\\
Before being fed to the CNN, the images were resized so that the shortest side was at 256 pixels and the center region of 224x224 pixels was cropped.

\section{Experimental results} \hypertarget{experimental_results}
In this Section we reported the results from the experiments we conducted on all the datasets described in \hyperlink{face_datasets}{Section 4}. We compared the results from the best two configurations, namely the ones in which we considered the state-of-the-art model with all parameters free to be trained first without and then with the curriculum learning and the teacher model supervision. The following nomenclature is used throughout the rest of the paper in order to indicate models that have been trained in each case: ``nT-nC" (without Teacher model nor Curriculum learning), ``T-C" (with Teacher model and Curriculum learning). Moreover, ``SotA model" will indicate the original state-of-the-art model.\\
In the latter configuration we used the loss function given in \autoref{eq:loss} for which we empirically discovered that the best value for the balancing weight $\lambda$ was 0.1. Specifically, we observed that the results for different values where consistent within less than 10\% and that the $\lambda = 0.1$ gave us the best results.\\
We tested our models on the 1:1 face verification and 1:N face identification protocols.\\
In the former case, the performances were evaluated through the Receiver Operating Characteristics (ROC) metric, which measures the true acceptance rate (TAR) as a function of the false acceptance rate (FAR).\\
In the latter case, depending on whether the probe identity belongs to the gallery or not, face identification is divided into close- or open-set, respectively.\\
In the close-set scenario, the performance of a model was evaluated by means of the Cumulative Match Characteristics (CMC) metric, which measures the percentage of probes identified within a given rank, while in the open-set case the performance was evaluated by the Decision Error Tradeoff (DET), which reports the false negative identification rate (FNIR) as a function of the false positive identification rate (FPIR).\\

\subsection{IJB-B}
For what concern the IJB-B~\cite{whitelam2017iarpa} dataset, we used it in order to test our models on the 1:1 face verification and 1:N face identification (open- and close-set) protocols.\\
\subsubsection{1:1 Face verification}
\begin{figure}[!h]
\includegraphics[width=\textwidth]{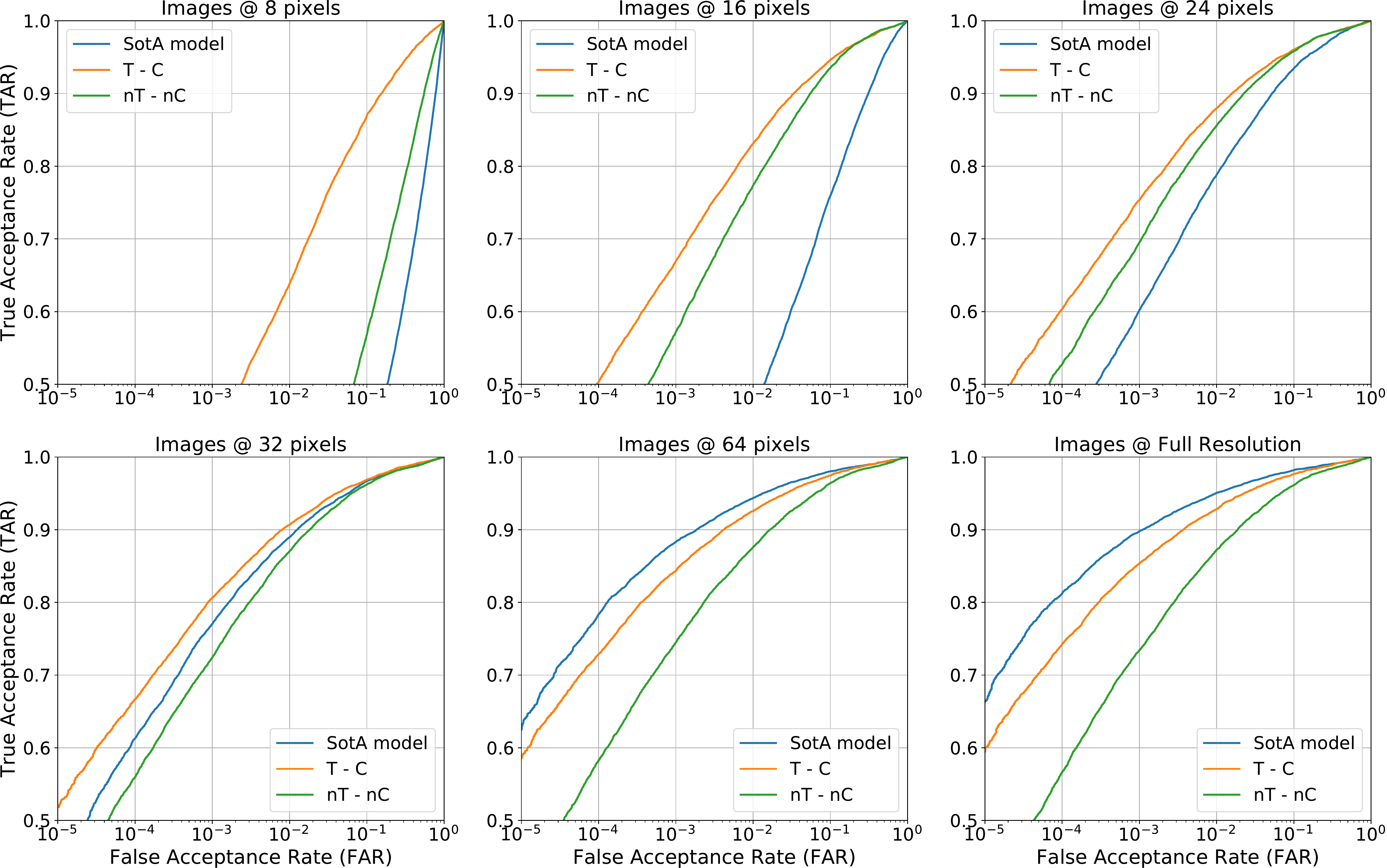}
\caption{ROC metric for SotA model model and fine-tuned ones. Each plot represent a different value of the input images resolution.} \label{fig:ijbb_roc}
\end{figure}

In order to evaluate the face verification protocol, a list of 8010270 comparisons is provided between two galleries templates, S1 and S2, and mixed media probe templates. In total, there are 10270 genuine comparisons and 8 millions impostor comparisons which allow to evaluate TAR at very low values of FAR which is important, for example, for surveillance-type applications.\\
The results are reported in \autoref{fig:ijbb_roc} in terms of the ROC metric for which we considered the images to be compared at the same resolution.

From \autoref{fig:ijbb_roc} we can notice that the model fine-tuned with our training algorithm reached the highest performances for resolution below 64 pixels. Interestingly, we can see that, on one side the use of the Teacher model helped to improve upon the performances at very LR and, on the other side, it helped to reduced the drop in performances at HR thus, enhancing the discriminative power of the model.\\
Even though at higher resolutions than 32 pixels the original model performed better than ours, the drop is only within few percent which is totally outweighed by the gain at LR. For a better comparison, in \autoref{tab:ijbb_tar_far_table} we reported the value of the TAR, at a reference FAR value of $1.e^{-3}$, for the various resolutions. 

\begin{table}[!h]
\caption{TAR values for a reference value of the FAR equals to $1.e^{-3}$. Each column represents a different value of the input image resolution. The last column stands for images that has been used at the original resolution.}
\label{tab:ijbb_tar_far_table}
\begin{tabularx}{\textwidth}{|c|>{\centering}X|>{\centering}X|>{\centering}X|>{\centering}X|>{\centering}X|>{\centering}X|>{\centering\arraybackslash}X|}
\hline
Model & \multicolumn{7}{c|}{TAR@FAR=$1.e^{-3}$ for different image resolution (pixels)} \\ 
      & 8 & 16 & 24 & 32 & 64 & 128 & full res  \\ \hline
SotA model&  0.048 &  0.244 &  0.602 &  0.770 &  \textbf{0.883} &  \textbf{0.895} &  \textbf{0.898} \\ \hline
nT-nC & 0.088 &  0.572 &  0.695 &  0.724 &  0.745 &  0.740 &  0.734 \\ \hline
T-C & \textbf{0.420} &  \textbf{0.670} &  \textbf{0.754} &  \textbf{0.807} &  0.843 &  0.851 &  0.854 \\ \hline
\end{tabularx}
\end{table}

From \autoref{tab:ijbb_tar_far_table} we can notice that, even though our best model experience a drop within $\sim$4\% at resolutions above 32 pixels, we have been able to improve the performance of a state-of-the-art model up to about one order of magnitude at LR.\\
We then considered the more challenging scenario of cross-resolution matches, which is perhaps the most interesting for forensic and surveillance systems applications. The results for the ``T-C" fine-tuned and the ``SotA model" models are reported in \autoref{tab:ijbb_cross_resolution_1}.

\begin{table*}[t]
\caption{True Acceptance Rate (TAR @ FAR = 1.e$^{-3}$) for cross-resolution face verification. Between brackets we reported the value from the original model.}\label{tab:ijbb_cross_resolution_1}
\begin{tabularx}{\textwidth}{|cc|>{\centering}X|>{\centering}X|>{\centering}X|>{\centering}X|>{\centering}X|>{\centering}X|>{\centering\arraybackslash}X|}
\hline
& & \multicolumn{7}{c|}{Resolution (pixel)} \\
& & 8 & 16 &  24 &  32 &  64 & 128 & 256 \\ \cline{2-9}
& 8 & 	\textbf{42.0}  (4.8) & & & & & &\\ \cline{2-9} 
& 16 & \textbf{41.8} (0.2) & \textbf{67.0} (24.4)& & & & &  \\ \cline{2-9} 
& 24 & \textbf{37.0} (0.2)& \textbf{69.7} (18.3)& \textbf{76.4} (60.2) & & & & \\ \cline{2-9} 
& 32 &  \textbf{32.1} (0.2)&\textbf{69.0} (9.0)& \textbf{76.6} (65.4) & \textbf{80.7} (77.0) & & & \\ \cline{2-9} 
& 64 & \textbf{28.0} (0.2)&\textbf{67.1} (2.9) & \textbf{76.3} (60.4)& \textbf{82.1} (80.5) & 84.3 (88.3) & & \\ \cline{2-9} 
& 128 &  \textbf{27.1} (0.1)&\textbf{66.3} (2.4) & \textbf{76.0} (57.9)& \textbf{82.1} (80.0) &  84.7 (88.8) & 85.1 (89.6) & \\ \cline{2-9} 
\multirow{-13}{*}{Resolution (pixel)   } & 256 & \textbf{27.2} (0.1) & \textbf{66.3} (2.3) & \textbf{75.9} (57.4)  & \textbf{82.2} (80.1)& 84.9 (89.0)& 85.2 (89.7)& 85.4 (89.8)   \\ \hline
\end{tabularx}
\end{table*}

As we can see from \autoref{tab:ijbb_cross_resolution_1}, even though our model is characterized by a performance which is up to $\sim$4\% below the one of the state-of-the-art model, such drop is completely outweighed by the improvements, up to about a factor 200, at resolutions below 64 pixels. We especially acknowledge the improvements at 8 and 16 pixels.

\subsubsection{1:N Face identification}
\begin{figure}[!h]
\begin{tabular}{c}
\includegraphics[width=\textwidth]{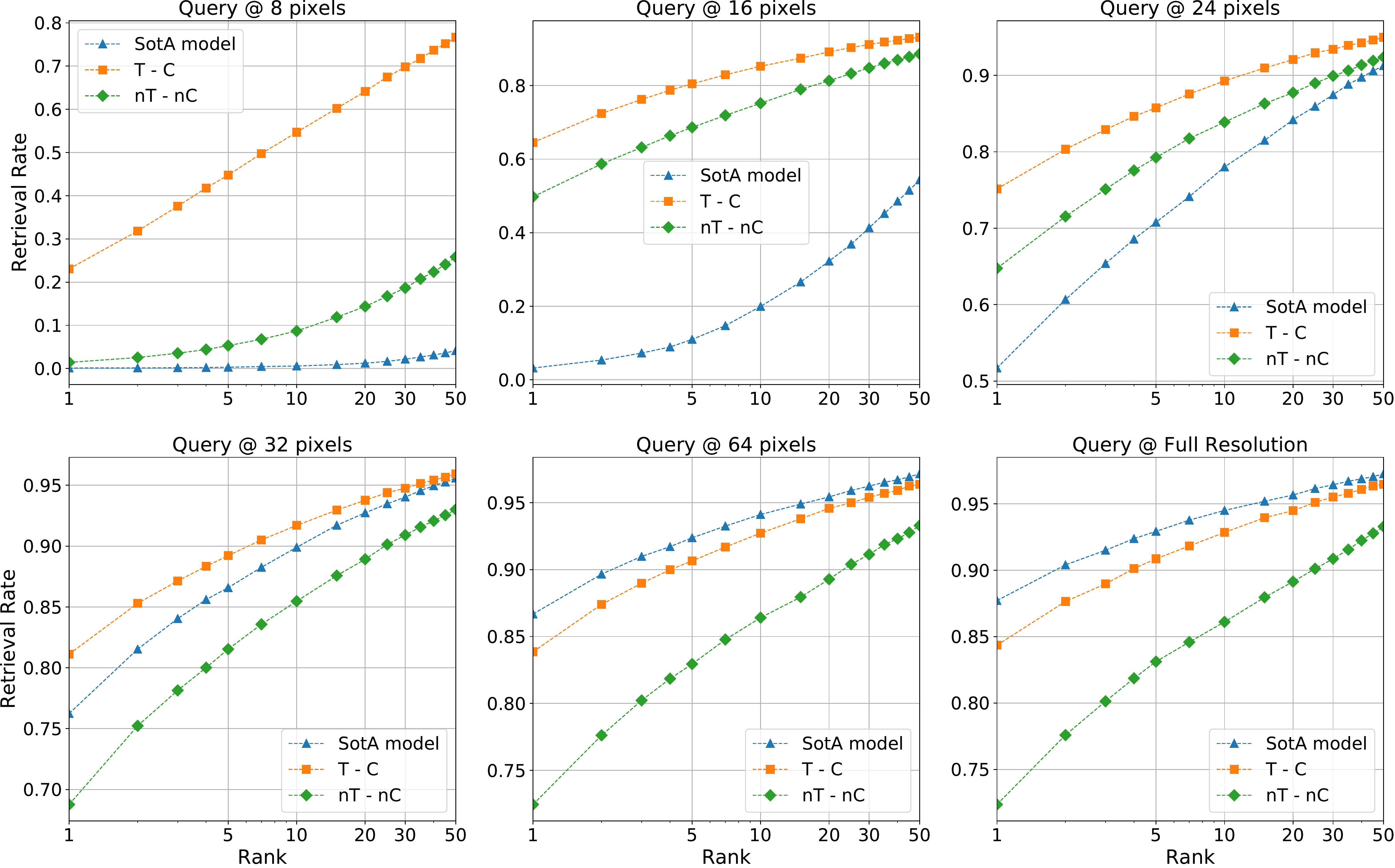}
\end{tabular}
\caption{CMC for SotA model and our fine-tuned ones. The higher the curve the better the discrimination power of the model. Each plot correspond to a different probe resolution.} \label{fig:ijbb_cmc}
\end{figure}

For the face identification protocol we considered the close- and open-set scenarios.\\
For the close-set scenario we measured the CMC which reports the fraction of searches that returned a true match within a rank \textit{r}. We considered two scenarios: one in which probe and gallery were at the same resolution, and one in which the probe was at different resolutions while the gallery was at full resolution. Since the latter case seems to be the more realistic one, we only show the corresponding results in \autoref{fig:ijbb_cmc}.

From \autoref{fig:ijbb_cmc} it is clear that up to 32 pixels, the performances of our models totally overcome the ones from the state-of-the-art model. To make this assessment more clear, in \autoref{tab:ijbb_cmc_table} we report the Rank-1 measurement, which is the most challenging one, obtained from the various model considering the various probe resolutions as in \autoref{fig:ijbb_cmc}.

\begin{figure}[!h]
\begin{tabular}{c}
\includegraphics[width=\textwidth]{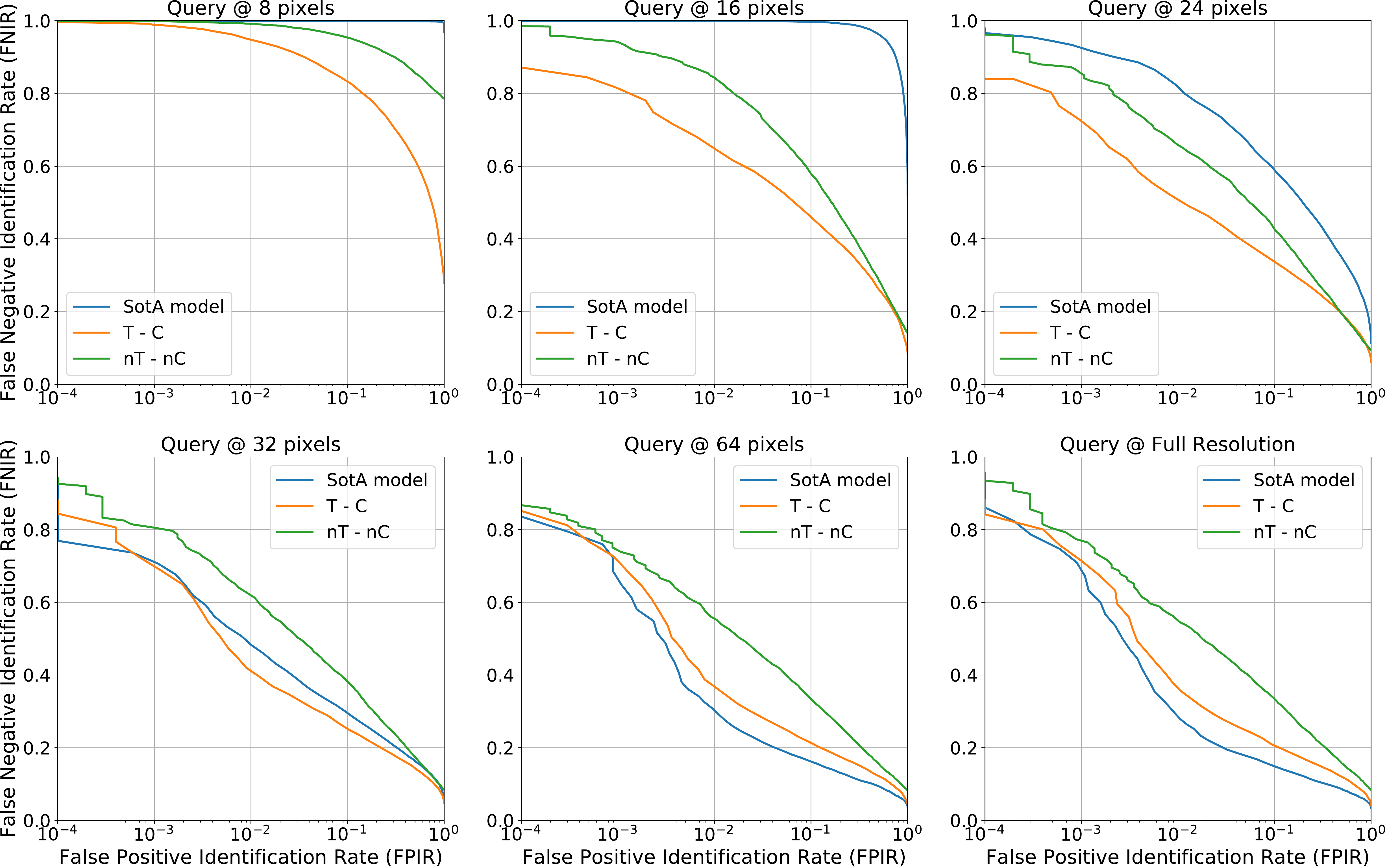}
\end{tabular}
\caption{DET for SotA model and our fine-tuned ones. The lower the curve the better the discrimination power of the model. Each plot correspond to a different probe resolution.} \label{fig:ijbb_det}
\end{figure}

\begin{table}[!h]
\caption{Rank-1 performance on IJB-B dataset, for the 1:N identification close-set protocol, considering probes at different resolutions.}
\label{tab:ijbb_cmc_table}
\begin{tabularx}{\textwidth}{|c|>{\centering}X|>{\centering}X|>{\centering}X|>{\centering}X|>{\centering}X|>{\centering}X|>{\centering\arraybackslash}X|}
\hline
 Model & \multicolumn{7}{c}{Rank-1 for different query resolution (pixels)} \\ 
                              &  8 & 16  & 24  & 32 & 64  & 128  & full res \\ \hline
SotA model & 0.001 &  0.032 &  0.517 &  0.762 &  \textbf{0.867} &  \textbf{0.877} &  \textbf{0.877}  \\ \hline
nT-nC     & 0.014 &  0.498 &  0.648 &  0.688 &  0.724 &  0.725 &  0.724 \\ \hline
T-C       & \textbf{0.231} &  \textbf{0.645} &  \textbf{0.751} &  \textbf{0.811} &  0.839 &  0.844 &  0.844 \\ \hline
\end{tabularx}
\end{table}
As we can see from \autoref{tab:ijbb_cmc_table}, even though our model has a rank-1 score lower at most of $\sim$4\% with respect to the state-of-the-art model, with our training procedure we have gained up to about two orders of magnitude in performances at resolution below 64 pixels.\\
After that, we considered the more realistic open-set protocol~\cite{grother2018ongoing}, for forensic and surveillance systems applications, which is often referred as the ``watch-list" identification scenario. As we did previously for the close-set case, we considered the case in which probe and gallery were at the same resolution, and the case in which the probe was at different resolutions while the gallery was at full resolution. Since the latter case seems to be the more realistic one, we only show the corresponding DET results in \autoref{fig:ijbb_det}.

From \autoref{fig:ijbb_det} one can notice than once again the model, trained with our full training algorithm, performs better than the state-of-the-art model for resolutions below 64 pixels. In \autoref{tab:ijbb_det_table} we reported the value of the FNIR at a reference value of the FPIR equals to $1.e^{-2}$, for the various resolution of the input probe.

\begin{table*}[ht]
\caption{FNIR @ FPIR = $1.e^{-2}$ for the tested models. The various columns correspond to a different resolution of the input probe.}\label{tab:ijbb_det_table}
\begin{tabularx}{\textwidth}{|c|>{\centering}X|>{\centering}X|>{\centering}X|>{\centering}X|>{\centering}X|>{\centering}X|>{\centering\arraybackslash}X|}
\hline
 Model & \multicolumn{7}{c}{FPIR@FNIR=$1.e-2$ for different query resolution (pixels)} \\ 
                              &  8 & 16  & 24  & 32 & 64  & 128  & full res \\ \hline
SotA model & 1.000 &  0.999 &  0.876 &  0.532 & \textbf{0.291} &  \textbf{0.269} &  \textbf{0.268} \\ \hline
nT-nC & 0.992 &  0.841 &  0.655 &  0.610 &  0.550 &  0.550 &  0.551 \\ \hline
T-C & \textbf{0.956} &  \textbf{0.654} & \textbf{0.505} & \textbf{0.412} &  0.368 &  0.366 &  0.364 \\ \hline
\end{tabularx}
\end{table*}

\subsection{IJB-C}
In this section we conducted on the IJB-C dataset the same experiments we did on the IJB-B one. Since the evaluated exactly the same metrics, we won't spent too much text in order to explain why we used the specific evaluation criteria.

\subsubsection{1:1 Face verification}

\begin{figure}[!h]
\begin{tabular}{c}
\includegraphics[width=\textwidth]{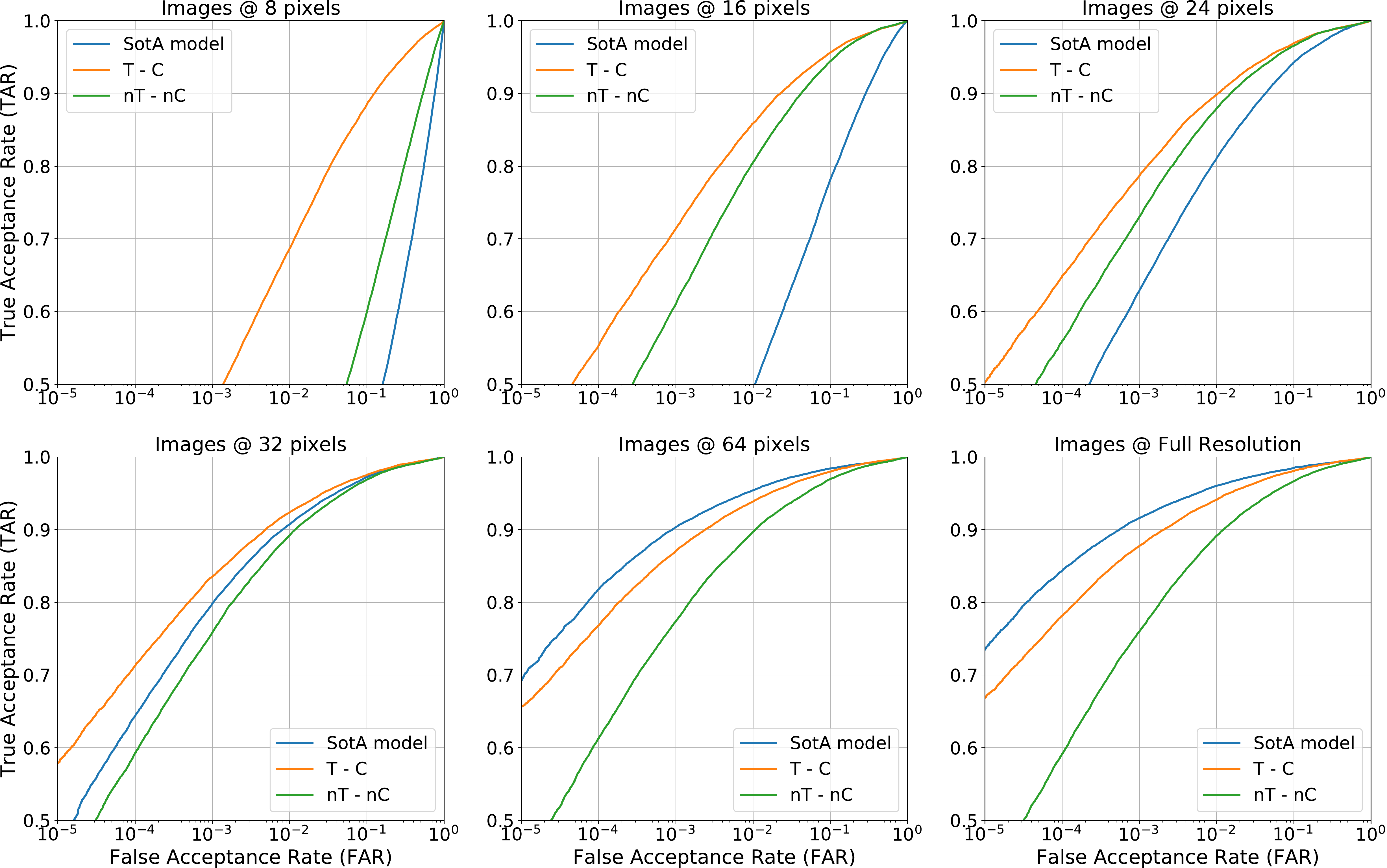}
\end{tabular}
\caption{ROC curves for SotA model model and fine-tuned ones. Each plot represent a different value of the input image resolution.} \label{fig:ijbc_roc}
\end{figure}

In order to evaluate the face verification protocol, a list
of about 15 millions comparisons is provided between two galleries templates, G1 and G2,
and mixed media probe templates. In total, there are about 20000  genuine comparisons
while all the remaining ones are impostor comparisons which allow to evaluate TAR at very low values of FAR that is important, for example, for surveillance-type applications.
The results are reported in \autoref{fig:ijbc_roc} in terms of the ROC metric for which we considered the images to be compared at the same resolution.

From \autoref{fig:ijbc_roc} they are evident the higher performances of our model when considering resolutions below 64 pixels. As we already saw previously, also in this case the use of the proposed training algorithm helped to improve upon the performances at LR and to mitigate the drop at HR. Indeed, even though at higher resolutions the original model performed better than ours, the drop is only within few percent which is totally outweighed by the gain at LR. For a better comparison, in \autoref{tab:ijbc_tar_far_table} we reported the value of the TAR, at a reference FAR value of $1.e^{-3}$, for the various resolutions.

\begin{table}[!h]
\caption{TAR values for a reference value of the FAR equals to $1.e^{-3}$. Each column represent a different value of the input image resolutions. The last column stands for images that has been used at the original resolution.}
\label{tab:ijbc_tar_far_table}
\begin{tabularx}{\textwidth}{|c|>{\centering}X|>{\centering}X|>{\centering}X|>{\centering}X|>{\centering}X|>{\centering}X|>{\centering\arraybackslash}X|}
\hline
Model & \multicolumn{7}{c|}{TAR @ FAR=$1.e^{-3}$ for different input image resolution (pixels)} \\ 
      & 8 & 16 & 24 & 32 & 64 & 128 & full res  \\ \hline
SotA model                & 0.052 &  0.262 &  0.629 &  0.798 &  \textbf{0.904} &  \textbf{0.914} &  \textbf{0.916} \\ \hline
nT-nC & 0.098 &  0.610 &  0.730 &  0.759 &  0.774 &  0.766 &  0.760 \\ \hline
T-C & \textbf{0.471} &  \textbf{0.714} &  \textbf{0.787} &  \textbf{0.835} &  0.871 &  0.876 &  0.878 \\ \hline
\end{tabularx}
\end{table}

From \autoref{tab:ijbc_tar_far_table} we can notice that, even though our models experience a drop within $\sim$4\% at resolutions from 64 pixels and above, we have been able to improve the performance of a state-of-the-art model up to about an order of magnitude at LR. We then considered the more challenging scenario of cross-resolution matches, which is perhaps the most interesting for forensic and surveillance systems applications. The results for the ``T-C" and ``SotA model" models are reported in \autoref{tab:ijbc_cross_resolution_1}.

\begin{table*}[ht]
\caption{True Acceptance Rate (TAR @ FAR = 1.e$^{-3}$) for cross-resolution face verification. Between brackets we reported the value from the original model.}\label{tab:ijbc_cross_resolution_1}
\begin{tabularx}{\textwidth}{|cc|>{\centering}X|>{\centering}X|>{\centering}X|>{\centering}X|>{\centering}X|>{\centering}X|>{\centering\arraybackslash}X|}
\hline
& & \multicolumn{7}{c|}{Resolution (pixel)} \\
& & 8 & 16 &  24 &  32 &  64 & 128 & 256 \\ \cline{2-9}
& 8 & 	\textbf{47.1}  (5.2) & & & & & &\\ \cline{2-9} 
& 16 & \textbf{46.4} (0.3) & \textbf{71.4} (26.2)& & & & &  \\ \cline{2-9} 
& 24 & \textbf{40.7} (0.2)& \textbf{73.7} (18.5) & \textbf{78.7} (62.9) & & & & \\ \cline{2-9} 
& 32 &  \textbf{34.7} (0.2)&\textbf{73.0} (9.7)& \textbf{79.8} (68.0) & \textbf{83.5} (79.8) & & & \\ \cline{2-9} 
& 64 & \textbf{30.0} (0.2)&\textbf{71.1} (3.8) & \textbf{79.6} (63.4)& \textbf{84.8} (83.0) & 87.1 (90.4) & & \\ \cline{2-9} 
& 128 &  \textbf{29.0} (0.1)&\textbf{70.3} (3.2) & \textbf{79.3} (60.9)& \textbf{84.9} (82.6) &  87.4 (90.8) &  87.6 (91.4) & \\ \cline{2-9} 
\multirow{-13}{*}{Resolution (pixel)   } & 256 & \textbf{29.0} (0.1) & \textbf{70.2} (3.1) & \textbf{79.2} (60.3)  & \textbf{84.9} (82.6) & 87.5 (90.9)& 87.7 (91.5) & 87.8  (91.6)   \\ \hline
\end{tabularx}
\end{table*}

As we can see from \autoref{tab:ijbc_cross_resolution_1} even though our model is characterized by a performance which up to $\sim$4\% below the one of the state-of-the-art model, such drop is completely outweighed by the improvements, up to about a factor 200, at resolutions below 64 pixels. We especially acknowledge the improvements at 8 and 16 pixels.

\subsubsection{1:N Face identification}

\begin{figure*}[ht]
\begin{tabular}{c}
\includegraphics[width=\textwidth]{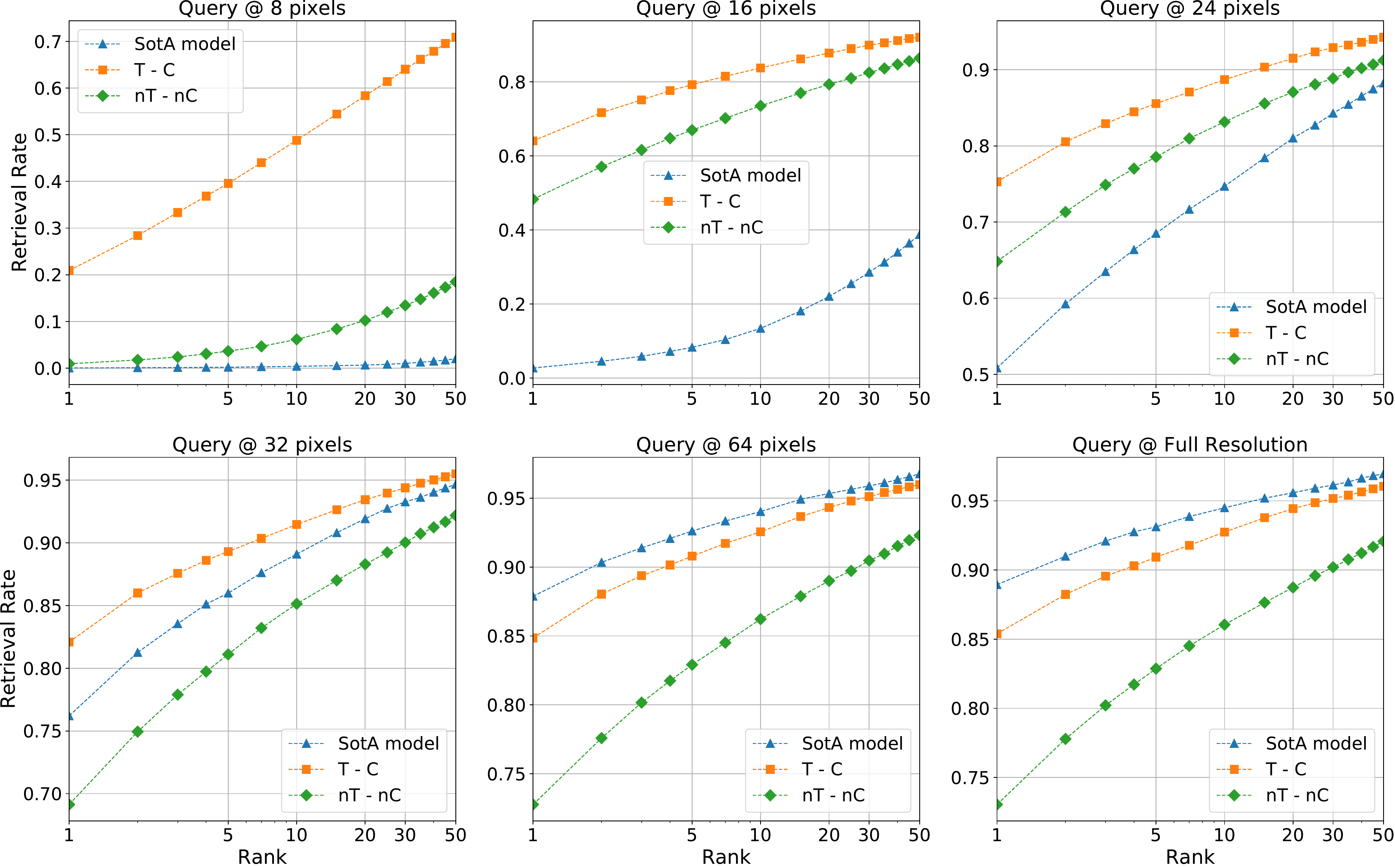}
\end{tabular}
\caption{CMC for SotA model and our fine-tuned ones. The higher the curve the better the discrimination power of the model. Each plot correspond to a different probe resolution.} \label{fig:ijbc_cmc}
\end{figure*}

As we did for the IJB-B~\cite{whitelam2017iarpa} dataset, also in this case we evaluated the CMC metric considering different probe resolutions while keeping the gallery at full resolution. The results are reported in \autoref{fig:ijbc_cmc}.

From \autoref{fig:ijbc_cmc} it is clear that up to 32 pixels, the performances of our model totally overcome the ones from the state-of-the-art model. To make this assessment more clear, in \autoref{tab:ijbc_cmc_table} we report the Rank-1 measurement, which is the most challenging one, obtained from the various model considering the various probe resolutions as in \autoref{fig:ijbc_cmc}.

\begin{table*}[ht]
\caption{Rank-1 performance on IJB-B dataset, for the 1:N identification close-set protocol, considering probes at different resolutions.}
\label{tab:ijbc_cmc_table}
\begin{tabularx}{\textwidth}{|c|>{\centering}X|>{\centering}X|>{\centering}X|>{\centering}X|>{\centering}X|>{\centering}X|>{\centering\arraybackslash}X|}
\hline
 Model & \multicolumn{7}{c}{Rank-1 for different query resolution (pixels)} \\ 
                    &  8 & 16  & 24  & 32 & 64  & 128  & full res \\ \hline
SotA model                   & 0.001 &  0.027 &  0.508 &  0.762 &  \textbf{0.879} &  \textbf{0.888} &  \textbf{0.889}  \\ \hline
nT-nC & 0.009 &  0.483 &  0.648 &  0.691 &  0.728 &  0.731 &  0.730 \\ \hline
T-C & \textbf{0.209} &  \textbf{0.641} &  \textbf{0.753} &  \textbf{0.821} &  0.849 &  0.854 &  0.854 \\ \hline
\end{tabularx}
\end{table*}

\begin{figure*}[ht]
\includegraphics[width=\textwidth]{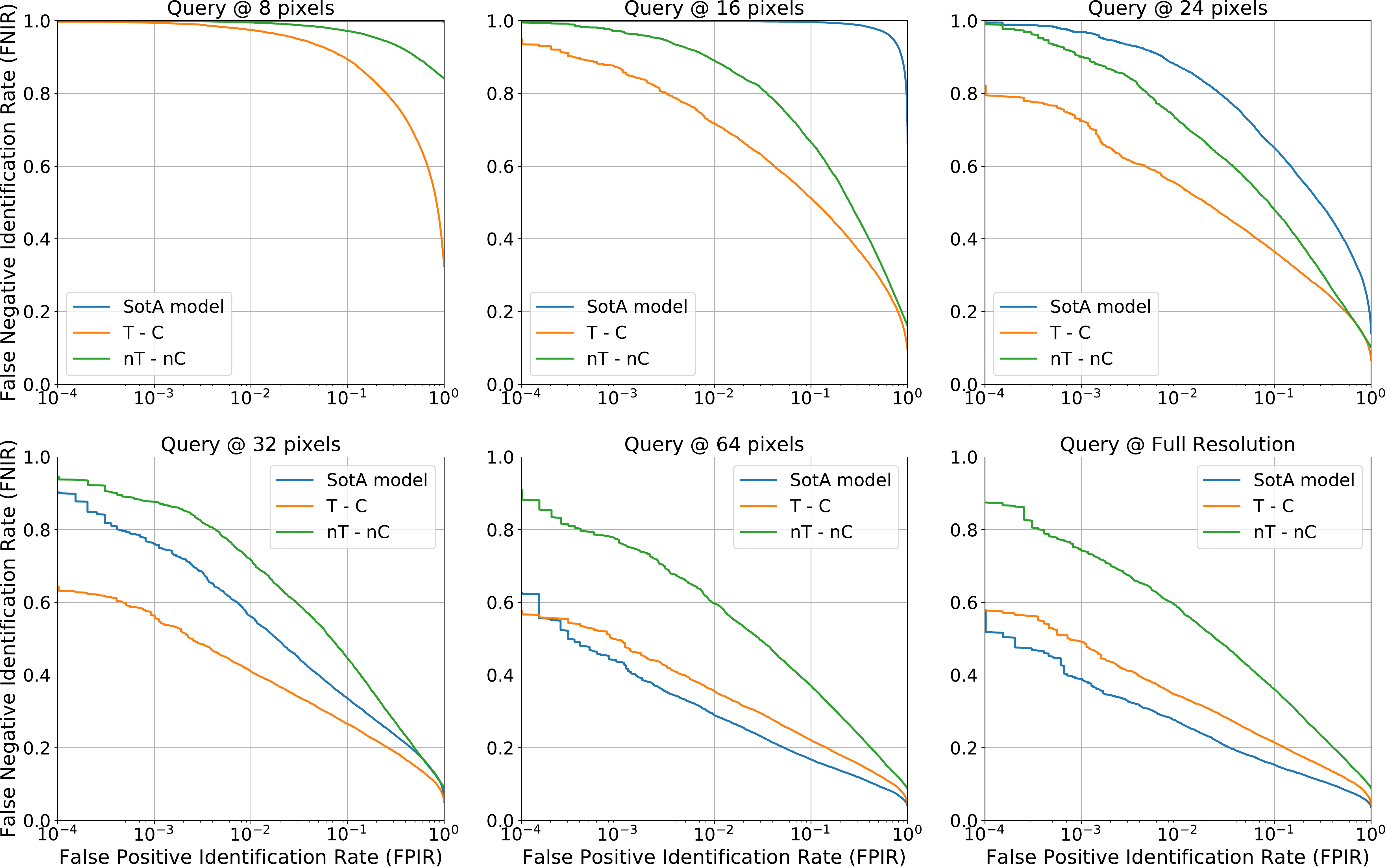}
\caption{DET for SotA model and our fine-tuned ones. The lower the curve the better the discrimination power of the model. Each plot correspond to a different probe resolution.} \label{fig:ijbc_det}
\end{figure*}

As we can see from \autoref{tab:ijbc_cmc_table}, even though our model has a rank-1 score lower at  most  of  $\sim$4\%  with  respect  to  the  state-of-the-art  model,  with  our  training procedure we have gained up to about two orders of magnitude in performances at resolution below 64 pixels. Then, we evaluated the DET metric and the results are shown in \autoref{fig:ijbc_det}.

From \autoref{fig:ijbc_det} one can notice than once again our models performs better than the state-of-the-art model for resolutions below 64 pixels. In \autoref{tab:ijbc_det_table} we reported the value of the FNIR at a reference value of the FPIR equals to $1.e^{-2}$, for the various resolution of the input probe.

\begin{table*}[hb]
\caption{FNIR @ FPIR = $1.e^{-2}$ for the tested models. The various columns correspond to a different resolution of the input probe.}\label{tab:ijbc_det_table}
\begin{tabularx}{\textwidth}{|c|>{\centering}X|>{\centering}X|>{\centering}X|>{\centering}X|>{\centering}X|>{\centering}X|>{\centering\arraybackslash}X|}
\hline
 Model & \multicolumn{7}{c}{FPIR@FNIR=$1.e^{-2}$ for different query resolution (pixels)} \\ 
                              &  8 & 16  & 24  & 32 & 64  & 128  & full res \\ \hline
SotA model                   & 1.000 &  0.999 &  0.876 &  0.532 &  \textbf{0.291} &  \textbf{0.269} & \textbf{0.268}  \\ \hline
nT-nC & .996 &  0.890 &  0.727 &  0.667 &  0.590 &  0.580 &  0.579 \\ \hline
T-C & \textbf{0.974} & \textbf{0.718} &  \textbf{0.549} &  \textbf{0.411} &  0.345 &  0.341 &  0.339 \\ \hline
\end{tabularx}
\end{table*}

\subsection{TinyFace}
\begin{figure*}[!ht]
\centering
\includegraphics[width=\textwidth]{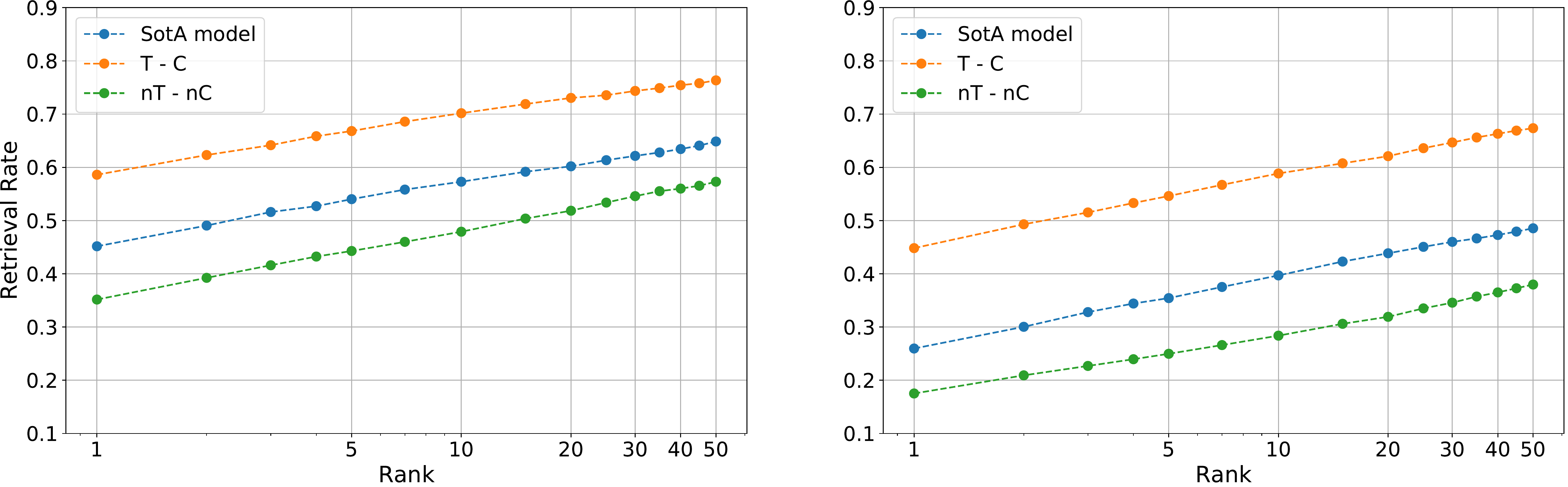}
\caption{Left: CMC for SotA model and fine-tuned ones. Right: same as Left but considering SR images.} \label{fig:tinyface_cmc}
\end{figure*}

On the TinyFace~\cite{cheng2018low} dataset we evaluated the CMC curve and the mean average precision (mAP) score. We compared our results with other state-of-the-art models, showing the higher performances of our models.\\
Since our final goal is to train a network to generate resolution-robust deep features that can be used for any task, we conducted our experiments on this dataset without fine-tuning our models on the faces contained in it. Instead, we only used the test set.\\
In \autoref{fig:tinyface_cmc}, we show the CMC, for the original state-of-the-art model alongside with ours. Moreover, we compared the effectiveness of our algorithm to the application of SR techniques for which we used the model from~\cite{kim2018deep}.


As we can see from \autoref{fig:tinyface_cmc}, our training procedure largely improves upon the performance of the pre-trained state-of-the-art model.  From the figures we observed a drop in performances when considering SR images. Since such behaviour is observed also in the state-of-the-art model, we can conclude that it is not due to our training algorithm. A possible explanation is that during the SR process, which is not optimized for discrimination but for producing high visual quality images, some identity information gets lost~\cite{ref_article5} thus, lowering the model discrimination power. In \autoref{tab:tinyface_rank} we report the retrieval rate and mAP score for the various models, while in \autoref{tab:tinyface_rank_sr} we report the same metric considering the SR input images. 

\begin{table*}[ht]
\caption{Retrieval rate and mAP score for various reference models from~\cite{cheng2018low} and for the ones under study in thi paper.}
\label{tab:tinyface_rank}
\begin{tabularx}{\textwidth}{|X|>{\centering}X|>{\centering}X|>{\centering}X|>{\centering\arraybackslash}X|}
\hline
Model                  & Rank-1 & Rank-20 & Rank-50 & mAP  \\ \hline \hline
DeepId2~\cite{cheng2018low}                & 17.4   & 25.2  & 28.3  & 12.1 \\ \hline
SphereFace~\cite{cheng2018low}             & 22.3   & 35.5  & 40.5  & 16.2 \\ \hline
VGGFace~\cite{cheng2018low}                & 30.4   & 40.4  & 42.7  & 23.1 \\ \hline
CentreFace~\cite{cheng2018low}             & 32.1   & 44.5  & 48.4  & 24.6 \\ \hline
SotA model      & 45.2 	& 60.2 	& 64.9  & 39.9 \\ \hline
nT-nC      & 46.0 & 63.5 & 68.0 & 41.7\\ \hline
T-C & \textbf{58.6} & \textbf{73.0} & \textbf{76.3} & \textbf{52.7} \\ \hline
\end{tabularx}
\end{table*}

\begin{table*}[ht]
\caption{Retrieval rate and mAP score, considering SR images, for various reference models from~\cite{cheng2018low} and for the ones under study in thi paper.}
\label{tab:tinyface_rank_sr}
\begin{tabularx}{\textwidth}{|X|>{\centering}X|>{\centering}X|>{\centering}X|>{\centering\arraybackslash}X|}
\hline
Model                  & Rank-1 & Rank-20 & Rank-50 & mAP  \\ \hline \hline
RPCN~\cite{cheng2018low} & 18.6 & 25.3 & 27.4 & 12.9 \\ \hline 
CSRI~\cite{cheng2018low} & 44.8 & 60.4 & 65.2 & 36.2 \\ \hline
SotA model      & 26.0 &	43.9 & 48.6 & 22.9 \\ \hline
nT-nC & 26.0 & 42.7 & 49.3 & 22.9 \\ \hline
T-C &  \textbf{44.8} & \textbf{62.1} & \textbf{67.4} & \textbf{39.7} \\ \hline
\end{tabularx}
\end{table*}

From \autoref{tab:tinyface_rank_sr} we can see that our models still performs better than others. Moreover, we would like to stress that the model with which we compare in \autoref{tab:tinyface_rank} and~\autoref{tab:tinyface_rank_sr} have been fine-tuned on the TinyFace~\cite{cheng2018low} dataset while ours did not. Such a result is critical since it means that our training procedure is able to train models characterized by a higher generalization capability.

\subsection{QMUL-SurvFace}
The QMUL-SurvFace~\cite{cheng2018surveillance} dataset is shipped with two protocols in order to test models: 1:1 face verification and 1:N open-set face identification.\\
To our knowledge this is the largest dataset containing real surveillance images. As we previously did for the Tinyface~\cite{cheng2018low}, also in this case we did not fine-tuned our model on the current dataset but we only used its test set.\\
Apart for the low resolution, this datasets is characterized by a high level of blurring and large head rotation angles, as it also clear from \autoref{fig:qmul_example}. These properties make the dataset very challenging for models which are not fine-tuned on it, as it will be clear from the results. Moreover, we would like to stress that for our work the most interesting cases were the cross-resolution matches thus, comparisons on only LR images is not our main aim.

\subsubsection{1:1 Face verification}
The face verification provides 5320 of positives and 5320 of negatives matches for which we evaluated the accuracy of the models. The results are reported in \autoref{tab:qmul_face_verification_acc_table}.

\begin{table*}[ht]
\caption{Accuracy on the 1:1 face verification protocol for the QMUL-SurvFace~\cite{cheng2018surveillance} dataset.}\label{tab:qmul_face_verification_acc_table}
\begin{center}
\begin{tabularx}{0.5\textwidth}{|>{\centering}X|>{\centering\arraybackslash}X|}
\hline
 Model & Accuracy (\%) \\  \hline \hline
SotA model & 56 \\ \hline
nT-nC & 69 \\ \hline
T-C & \textbf{72} \\ \hline
\end{tabularx}
\end{center}
\end{table*}

Then, we compared the effectiveness of our training method with SR methods. Specifically, we used the SR model from~\cite{kim2018deep}. The results are reported in \autoref{tab:qmul_face_verification_acc_table_sr}.

\begin{table*}[ht]
\caption{Accuracy on the 1:1 face verification protocol for the QMUL-SurvFace~\cite{cheng2018surveillance} dataset using deep features extracted from super resolved images.}\label{tab:qmul_face_verification_acc_table_sr}
\begin{center}
\begin{tabularx}{0.5\textwidth}{|>{\centering}X|>{\centering\arraybackslash}X|}
\hline
 Model & Accuracy (\%) \\  \hline \hline
SotA model & 55 \\ \hline
nT-nC & 60 \\ \hline
T-C & \textbf{61} \\ \hline
\end{tabularx}
\end{center}
\end{table*}

From \autoref{tab:qmul_face_verification_acc_table_sr} we see that the performances from our model decrease when tested on SR images, which is the same behaviour we already observed on the TinyFace~\cite{cheng2018low}. Thus, same considerations as before holds.

\subsubsection{1:N Face identification}
\begin{figure*}[ht]
\centering
\includegraphics[width=\textwidth]{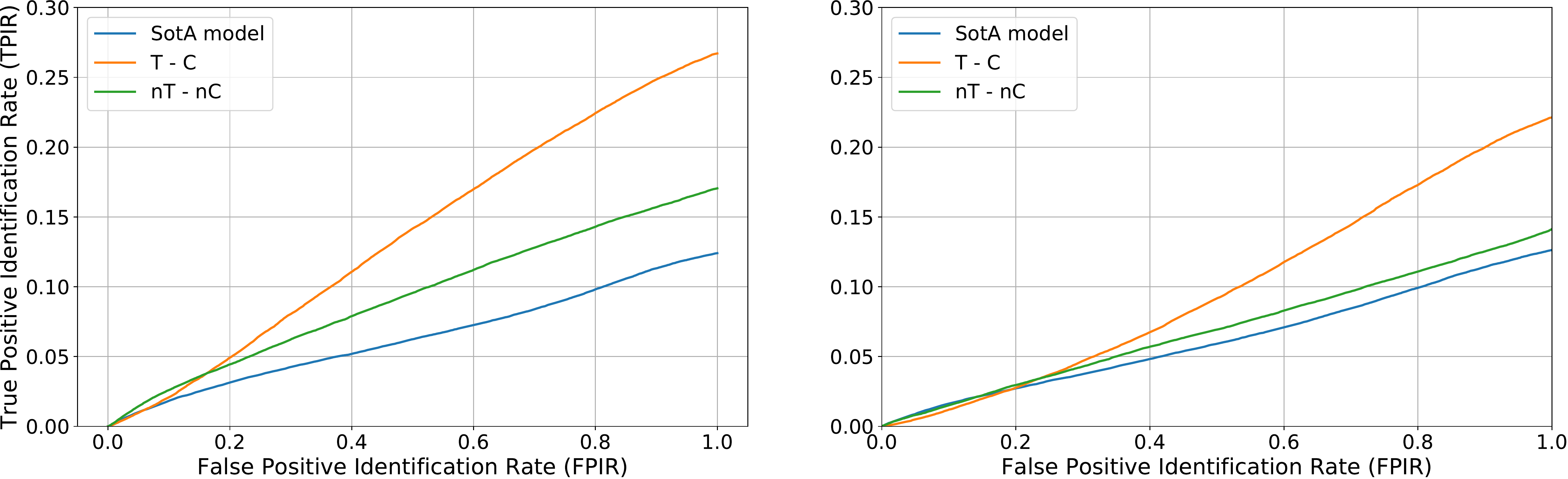}
\caption{True Positive Identification Rate (TPIR) as a function of the FPIR for SotA model and our fine-tuned ones. The higher the curve the better the discrimination power of the model. Left: original images. Right: SR images.} \label{fig:qmul_det}
\end{figure*}

The face identification protocol comprises the open-set scenario that for surveillance systems is of much more interest compared to the close-set one. \\
In order to be comparable with previously published results, in \autoref{fig:qmul_det} we report the True Positive Identification Rate (TPIR), which corresponds to 1-FNIR, as a function of the FPIR.

As we can see from \autoref{fig:qmul_det} our model obtained the best results. Moreover, we noticed a general drop in performances on SR images. As a summary statistics for the previous figures, in \autoref{tab:qmul_tpir_fpir_table} we report the values of area under the curve (AUC). Being available results from other models, we compared our results with others from~\cite{cheng2018surveillance}.


\begin{table*}[ht]
\caption{AUC values for different models. Reference models have been taken from~\cite{cheng2018surveillance}. In the last three rows, the value between brackets are the AUC obtained by using the SR images.}\label{tab:qmul_tpir_fpir_table}
\begin{center}
\begin{tabularx}{0.5\textwidth}{|>{\centering}X|>{\centering\arraybackslash}X|}
\hline
 Model & AUC (\%) \\ \hline \hline
DeepId2     & 7.9 \\ \hline
Centreface  & 7.6 \\ \hline 
FaceNet     & 6.4 \\ \hline 
SphereFace  & 9.0 \\ \hline
SotA model  & 6.4 (6.2) \\ \hline
nT-nC       & 9.3 (7.0) \\ \hline
T-C         & \textbf{13.8} (9.9) \\ \hline
\end{tabularx}
\end{center}
\end{table*}

From \autoref{tab:qmul_tpir_fpir_table} we can conclude that our CNN, trained with the proposed training algorithm, outperforms all previous state-of-the-art models on this dataset, without considering any specific fine-tuning on the QMUL-SurvFace~\cite{cheng2018surveillance} dataset. 

\section{Conclusion} \hypertarget{conclusion}
In this paper we focused on the FR task in the cross-resolution scenario. Specifically, we proposed a training approach in order to improve upon a state-of-the-art DL model ability to generate resolution-robust, up to a certain level, deep representations.\\
Nevertheless, even though we focused on the cross-resolution task, we noticed that with our training method we also obtained a noticeable improvement on FR among images at the same, but low, resolution i.e. down to 8 pixels.\\
In order to assess the effectiveness of our training algorithm, we tested two different fine-tuning configurations and compared the results with state-of-the-art models. We first considered images down sampled at a different resolutions in the range of [8, 256] pixels with a fixed frequency. Then, we used our proposed training algorithm in which we add a teacher signal and a variable down sampling frequency. We extensively tested all our models against the 1:1 face verification and the 1:N face identification protocols on various face datasets: IJB-B, IJB-C, QMUL-Survface and TinyFace.\\
On the first two datasets, the proposed approach gave us the best results on both low- and cross-resolution domains. Indeed, even though we observed a drop within $\sim$4\% percent of our model performance at resolutions strictly higher than 32 pixels, the improvement in the LR regime totally outweighed it. Specifically, we observed improvements up to about two orders of magnitude.\\
On the TinyFace, we set the new state-of-the-art considering the face identification close-set protocol which is shipped with the dataset.\\
On the QMUL-Survface our model performed better than any other state-of-the-art model which was not fine-tuned on the training split given by the dataset itself.\\
Moreover, we considered a super resolution approach for the previous two datasets. Interestingly, our measurements pointed out that even though our model still performed better than others, its performances were lower compared to when we used the original LR images. This could be a result of the fact that SR techniques are not optimized for discrimination thus, identity informations might be lost while super resolving the images. This might make our work more relevant for the scientific community.\\

\section*{Acknowledgments} This work was partially supported by the AI4EU project, funded by the EC (H2020 - Contract n. 825619). We gratefully acknowledge the support of NVIDIA Corporation with the donation of the Titan V GPU used for this research.



\begin{thebibliography}{10}
\expandafter\ifx\csname url\endcsname\relax
  \def\url#1{\texttt{#1}}\fi
\expandafter\ifx\csname urlprefix\endcsname\relax\def\urlprefix{URL }\fi
\expandafter\ifx\csname href\endcsname\relax
  \def\href#1#2{#2} \def\path#1{#1}\fi

\bibitem{ref_article1}
Y.~Wen, K.~Zhang, Z.~Li, Y.~Qiao, A discriminative feature learning approach
  for deep face recognition, in: European conference on computer vision,
  Springer, 2016, pp. 499--515.

\bibitem{ref_article2}
F.~Wang, J.~Cheng, W.~Liu, H.~Liu, Additive margin softmax for face
  verification, IEEE Signal Processing Letters 25~(7) (2018) 926--930.

\bibitem{turk1991face}
M.~A. Turk, A.~P. Pentland, Face recognition using eigenfaces, in: Proceedings.
  1991 IEEE Computer Society Conference on Computer Vision and Pattern
  Recognition, IEEE, 1991, pp. 586--591.

\bibitem{liu2002gabor}
C.~Liu, H.~Wechsler, Gabor feature based classification using the enhanced
  fisher linear discriminant model for face recognition, IEEE Transactions on
  Image processing 11~(4) (2002) 467--476.

\bibitem{ahonen2006face}
T.~Ahonen, A.~Hadid, M.~Pietikainen, Face description with local binary
  patterns: Application to face recognition, IEEE Transactions on Pattern
  Analysis \& Machine Intelligence~(12) (2006) 2037--2041.

\bibitem{wang2018deep}
M.~Wang, W.~Deng, Deep face recognition: A survey, arXiv preprint
  arXiv:1804.06655.

\bibitem{ref_article4}
W.~W. Zou, P.~C. Yuen, Very low resolution face recognition problem, IEEE
  Transactions on image processing 21~(1) (2012) 327--340.

\bibitem{cao2018vggface2}
Q.~Cao, L.~Shen, W.~Xie, O.~M. Parkhi, A.~Zisserman, Vggface2: A dataset for
  recognising faces across pose and age, in: 2018 13th IEEE International
  Conference on Automatic Face \& Gesture Recognition (FG 2018), IEEE, 2018,
  pp. 67--74.

\bibitem{ref_article5}
K.~Zhang, Z.~Zhang, C.-W. Cheng, W.~H. Hsu, Y.~Qiao, W.~Liu, T.~Zhang,
  Super-identity convolutional neural network for face hallucination, in:
  Proceedings of the European Conference on Computer Vision (ECCV), 2018, pp.
  183--198.

\bibitem{ref_article6}
X.~Yu, B.~Fernando, B.~Ghanem, F.~Porikli, R.~Hartley, Face super-resolution
  guided by facial component heatmaps, in: Proceedings of the European
  Conference on Computer Vision (ECCV), 2018, pp. 217--233.

\bibitem{ref_article7}
M.~Jian, K.-M. Lam, Simultaneous hallucination and recognition of
  low-resolution faces based on singular value decomposition, IEEE Transactions
  on Circuits and Systems for Video Technology 25~(11) (2015) 1761--1772.

\bibitem{ekenel2005multiresolution}
H.~K. Ekenel, B.~Sankur, Multiresolution face recognition, Image and Vision
  Computing 23~(5) (2005) 469--477.

\bibitem{luo2019multi}
X.~Luo, Y.~Xu, J.~Yang, Multi-resolution dictionary learning for face
  recognition, Pattern Recognition 93 (2019) 283--292.

\bibitem{he2016deep}
K.~He, X.~Zhang, S.~Ren, J.~Sun, Deep residual learning for image recognition,
  in: Proceedings of the IEEE conference on computer vision and pattern
  recognition, 2016, pp. 770--778.

\bibitem{hu2017squeeze}
J.~Hu, L.~Shen, G.~Sun, Squeeze-and-excitation networks. arxiv (2017).

\bibitem{whitelam2017iarpa}
C.~Whitelam, E.~Taborsky, A.~Blanton, B.~Maze, J.~Adams, T.~Miller, N.~Kalka,
  A.~K. Jain, J.~A. Duncan, K.~Allen, et~al., Iarpa janus benchmark-b face
  dataset, in: Proceedings of the IEEE Conference on Computer Vision and
  Pattern Recognition Workshops, 2017, pp. 90--98.

\bibitem{maze2018iarpa}
B.~Maze, J.~Adams, J.~A. Duncan, N.~Kalka, T.~Miller, C.~Otto, A.~K. Jain,
  W.~T. Niggel, J.~Anderson, J.~Cheney, et~al., Iarpa janus benchmark-c: Face
  dataset and protocol, in: 2018 International Conference on Biometrics (ICB),
  IEEE, 2018, pp. 158--165.

\bibitem{cheng2018surveillance}
Z.~Cheng, X.~Zhu, S.~Gong, Surveillance face recognition challenge, arXiv
  preprint arXiv:1804.09691.

\bibitem{cheng2018low}
Z.~Cheng, X.~Zhu, S.~Gong, Low-resolution face recognition, in: Asian
  Conference on Computer Vision, Springer, 2018, pp. 605--621.

\bibitem{guo2016ms}
Y.~Guo, L.~Zhang, Y.~Hu, X.~He, J.~Gao, Ms-celeb-1m: A dataset and benchmark
  for large-scale face recognition, in: European Conference on Computer Vision,
  Springer, 2016, pp. 87--102.

\bibitem{parkhi2015deep}
O.~M. Parkhi, A.~Vedaldi, A.~Zisserman, et~al., Deep face recognition., in:
  bmvc, Vol.~1, 2015, p.~6.

\bibitem{gunther2017unconstrained}
M.~G{\"u}nther, P.~Hu, C.~Herrmann, C.-H. Chan, M.~Jiang, S.~Yang, A.~R.
  Dhamija, D.~Ramanan, J.~Beyerer, J.~Kittler, et~al., Unconstrained face
  detection and open-set face recognition challenge, in: 2017 IEEE
  International Joint Conference on Biometrics (IJCB), IEEE, 2017, pp.
  697--706.

\bibitem{grgic2011scface}
M.~Grgic, K.~Delac, S.~Grgic, Scface--surveillance cameras face database,
  Multimedia tools and applications 51~(3) (2011) 863--879.

\bibitem{zou2011very}
W.~W. Zou, P.~C. Yuen, Very low resolution face recognition problem, IEEE
  Transactions on image processing 21~(1) (2011) 327--340.

\bibitem{tb}
E.~Zangeneh, M.~Rahmati, Y.~Mohsenzadeh, Low resolution face recognition using
  a two-branch deep convolutional neural network architecture, arXiv preprint
  arXiv:1706.06247.

\bibitem{sb}
S.~Shekhar, V.~M. Patel, R.~Chellappa, Synthesis-based robust low resolution
  face recognition, arXiv preprint arXiv:1707.02733.

\bibitem{lu2018deep}
Z.~Lu, X.~Jiang, A.~Kot, Deep coupled resnet for low-resolution face
  recognition, IEEE Signal Processing Letters 25~(4) (2018) 526--530.

\bibitem{iarpa}
{IARPA's Janus}, \url{https://www.iarpa.gov/index.php/research-programs/janus}.

\bibitem{bengio2009curriculum}
Y.~Bengio, J.~Louradour, R.~Collobert, J.~Weston, Curriculum learning, in:
  Proceedings of the 26th annual international conference on machine learning,
  ACM, 2009, pp. 41--48.

\bibitem{hinton2015distilling}
G.~Hinton, O.~Vinyals, J.~Dean, Distilling the knowledge in a neural network,
  arXiv preprint arXiv:1503.02531.

\bibitem{grother2018ongoing}
P.~J. Grother, M.~L. Ngan, K.~K. Hanaoka, Ongoing face recognition vendor test
  (frvt) part 2: Identification, Tech. rep. (2018).

\bibitem{kim2018deep}
J.-H. Kim, J.-S. Lee, Deep residual network with enhanced upscaling module for
  super-resolution., in: Proceedings of the IEEE Conference on Computer Vision
  and Pattern Recognition (CVPR) Workshops, 2018.

\end{thebibliography}
\end{document}